\documentclass[10pt,twocolumn,letterpaper]{article}
\usepackage[pagenumbers]{cvpr}
\usepackage{float,capt-of,algorithm,algpseudocode}

\providecommand{\mainbudget}{30K}

\providecommand{\tightlist}{\setlength{\itemsep}{0pt}\setlength{\parskip}{0pt}}

\definecolor{cvprblue}{rgb}{0.21,0.49,0.74}

\usepackage[breaklinks,colorlinks,allcolors=cvprblue]{hyperref}
\title{Streaming Visual Memory for Generative World-Action Models}
\author{%
Shizhuo Zhang$^{1,*}$ \quad Jiajun Wen$^{2,*}$ \quad Yuanpeng He$^{3,\dagger}$ \quad Hao Tang$^{4,\dagger}$ \quad Fan He$^{4}$ \\
Liang Chai$^{1}$ \quad Hao Wang$^{5,6}$ \quad Zhiliang Chen$^{6,7}$ \quad Yifan Zhao$^{1}$ \quad Yuchen Liu$^{1}$ \\
Huixin Zhang$^{4}$ \quad Jianhua Han$^{4}$ \quad Hang Xu$^{4}$ \quad Xiaodan Liang$^{5}$ \\[3pt]
\small $^{1}$Shanghai Jiao Tong University \quad $^{2}$Tsinghua University \quad $^{3}$Peking University \\
\small $^{4}$Yinwang Intelligent Technology Co. Ltd. \enspace $^{5}$Sun Yat-sen University \enspace $^{6}$Beijing Zhongguancun Academy \enspace $^{7}$Fudan University}

\begin{document}
\twocolumn[{%
\renewcommand\twocolumn[1][]{#1}%
\maketitle
\begin{center}
\captionsetup{type=figure}
\includegraphics[width=\textwidth]{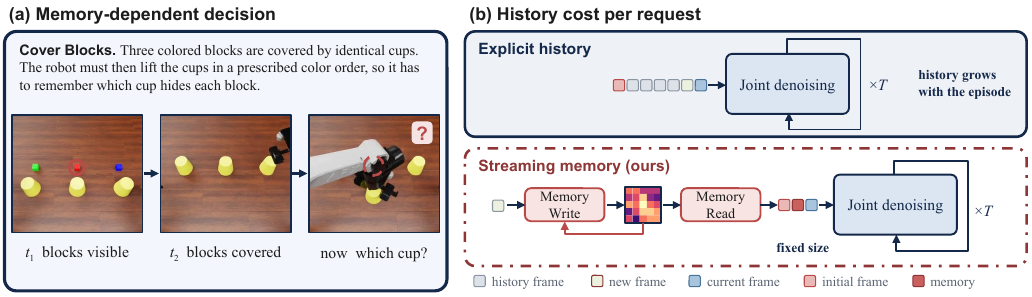}
\captionof{figure}{\textbf{Streaming visual memory for generative WAMs.} (a) In Cover Blocks, the cups look identical once the blocks are hidden, so the correct action depends on which block was visible earlier. At the current request, the arm reaches for the cup that covered the red block. Frames are head-camera crops from a closed-loop Streaming-history episode. (b) Explicit history grows with the episode, and every denoising iteration processes all sampled history frames. Our memory writes only the newly sampled frames into a fixed-size state and reads it once per request. All $T$ denoising iterations then reuse a fixed-size history condition of initial-frame and memory tokens.}
\label{fig:teaser}
\end{center}\vspace{0.6em}}]

\begingroup
\renewcommand{\thefootnote}{}
\footnotetext{\hspace{-1em}$^{*}$Equal contribution. \quad $^{\dagger}$Corresponding authors.}
\endgroup
\begin{abstract}
Generative world-action models (WAMs) jointly denoise future video and robot actions, yet manipulation often depends on visual evidence that is no longer visible. Explicitly conditioning on past frames increases the context processed at every denoising iteration. We introduce streaming visual memory for generative WAMs, which pairs a compressed visual memory with an efficient streaming procedure for maintaining and using it. Sampled multi-view observations are written patch by patch into a fixed-size memory state with a gated delta rule. The current observation extracts features from the strictly-past state, and the readout enters the video-side condition as memory tokens alongside initial-frame tokens, leaving the backbone unchanged. Across action requests, the state incorporates only newly sampled frames. Within a request, one readout is reused in every denoising iteration. With EWAM as the backbone, we compare this design with No-history, Window-history, and Full-history on six memory-dependent RoboDojo tasks using checkpoints trained for 30K updates. Streaming-history achieves 21.3\% success, compared with 7.7\%, 18.3\%, and 16.7\% for No-history, Window-history, and Full-history, respectively, while representing history with a fixed-size state and conditioning sequence. The largest gains occur on Cover Blocks, where success depends on recalling evidence that is no longer visible. On one RTX 5090 with the same rotary-position caching applied to all methods, Streaming-history has a median request latency of 1.119\,s, 25\% lower than Window-history and a $3.9\times$ paired per-request speedup over Full-history.
\end{abstract}

\section{Introduction}
\label{sec:intro}

Generative world-action models (WAMs) predict future visual observations and robot actions jointly, so that visual foresight learned from video can shape action generation~\cite{wang2026ewam}. Visuomotor policies and vision-language-action (VLA) models commonly condition on the current observation or a short observation window~\cite{kim2024openvla,black2026pi0visionlanguageactionflowmodel}. Tasks that depend on evidence outside this context require a representation of the past~\cite{kaelbling1998planning}. In Cover Blocks, the robot must uncover blocks in a prescribed color order after they have been hidden under identical cups. In Imitate Sorting Sequence, it must reproduce an order demonstrated earlier in the episode. Two moments can look the same yet require different actions (\cref{fig:teaser}a). Memory benchmarks for manipulation make this requirement explicit~\cite{fang2025sam2act,chen2026rmbench}, and the RoboDojo study observes that world-model-style prediction does not necessarily preserve or retrieve such sparse evidence after it disappears~\cite{chen2026robodojo}.

Supplying history to a generative WAM is costly in a specific way. The simplest approach adds past frames as further visual conditions. A bounded window keeps the context small but loses events that leave it unless they appear in the initial frame~\cite{guo2026video}. Retaining all sampled frames preserves access to the episode but enlarges the context processed by the backbone. In EWAM~\cite{wang2026ewam}, visual conditions take part in every iteration of joint video-action denoising, so explicit history is paid for at each iteration rather than once per request, and caching past visual encodings does not remove this cost. Hybrid historical caches in MemoryWAM and event memories in DiM-WAM offer alternative ways to manage this trade-off~\cite{yang2026memorywam,wang2026dim}.

These costs motivate a compact memory whose conditioning length stays fixed as history grows. The way this memory is executed also matters. Between action requests, only the newly sampled history frames need to be incorporated. Within a request, the observations remain fixed while future video and actions are refined. A persistent state can therefore accumulate history incrementally and supply one current-conditioned readout throughout denoising (\cref{fig:teaser}b).

We develop streaming visual memory for generative WAMs, which pairs a compressed visual memory with an efficient streaming inference procedure. Sampled multi-view observations are encoded as patch features and written into a fixed-size memory state with a gated delta rule~\cite{schlag2021linear,yang2025gated}. A frame-level gate controls retention, and ordered patch updates correct the state's prediction of each key-value association. Patches of the current observation query the strictly-past state by content and image-plane position, extracting one historical feature per patch. The readout enters the video-side condition of the WAM as memory tokens alongside the tokens of an initial frame and the current observation. Because the module only consumes patch tokens and returns condition tokens, it leaves the backbone unchanged and applies to WAMs that generate future video from visual condition tokens. In EWAM, history then reaches the Action Expert through the existing asymmetric video-action attention. At deployment, incremental updates maintain the state across requests, and a single readout supplies all denoising iterations within each request.

We evaluate this design on the six memory tasks of RoboDojo~\cite{chen2026robodojo}, comparing four history representations at a common budget and isolating the cost of memory maintenance with a same-checkpoint prefix-reconstruction control. All closed-loop comparisons use checkpoints trained for 30K updates with the same backbone, data split, global batch size, and evaluation layouts. Streaming-history succeeds in 64 of 300 episodes, compared with 23 for No-history, 55 for Window-history, and 50 for Full-history. Matched serving measurements on one RTX 5090 give a median request latency of 1.119\,s, a 25\% reduction relative to Window-history and a $3.9\times$ speedup over Full-history.

Using EWAM as the backbone, we make three contributions.
\begin{enumerate}
\tightlist
\item \textbf{Compressed visual memory for generative WAMs.} We represent sampled multi-view history with a fixed-size, patch-level memory state and extract current-conditioned features for joint video-action generation. The module attaches to the video-side condition by concatenating the readout with an initial frame and the current observation, without modifying the backbone.
\item \textbf{Efficient streaming inference.} We maintain memory through incremental updates across action requests and reuse one readout in every denoising iteration of a request, reducing repeated history processing while keeping the conditioning length fixed.
\item \textbf{Controlled evaluation of history and inference cost.} With checkpoints trained for 30K updates, we compare four history representations on six memory-dependent manipulation tasks. We separately measure inference cost under controlled serving configurations. A same-checkpoint prefix-reconstruction control isolates the benefit of incremental updates.
\end{enumerate}

\section{Related Work}
\label{sec:related}

\subsection{Memory in Robot Policies}
\label{memory-in-robot-policies}
Memory-dependent manipulation has motivated dedicated benchmarks~\cite{fang2025sam2act,chen2026rmbench}, including the Memory dimension of RoboDojo used in this work~\cite{chen2026robodojo}. A common approach supplies past observations explicitly. ContextVLA amortizes a multi-frame context into one token~\cite{jang2025contextvla}, HAMLET attends over cached per-step moment tokens~\cite{koo2026hamlet}, and keyframe-based policies select and retain sparse past views~\cite{sridhar2026scaling}. Studies of causal confusion and copycat shortcuts further show that the usefulness of longer histories depends on how policies learn to use them~\cite{de2019causal,wen2020fighting}.

Other policies learn a compact memory. MemoryVLA retrieves and fuses perceptual and cognitive memories before action decoding~\cite{shi2026memoryvla}, ReCAT combines a Mamba-2 state with a causal-attention cache~\cite{vanjani2026recat}, and RoboTTT compresses long contexts into test-time-trained fast weights~\cite{jiang2026robottt}.

Associative memory with delta-rule updates maintains a fixed-size key-value state~\cite{schlag2021linear,yang2024parallelizing,yang2025gated}. Dual-Stream Memory combines such a temporal summary with task-conditioned historical retrieval~\cite{wang2026dual}. DRAM writes each frame jointly into the state~\cite{zhao2026dram}, and MemBodied stores action-observation transitions alongside an initial-scene anchor~\cite{pala2026membodied}. These memories condition an action policy. We study patch-level visual memory as a condition of joint video-action generation, together with its maintenance across action requests.

\subsection{History and Computation Reuse in Generative WAMs}
\label{historical-conditioning-in-generative-world-action-models}
Video-prediction policies and world-action models couple visual prediction with action generation~\cite{du2023learning,hu2024video,ye2026world}. Motus and EWAM jointly denoise video and actions with video, vision-language, and action experts~\cite{bi2026motus,wang2026ewam}. These models differ in how they retain history. Many condition on the current frame or a short window~\cite{hu2024video,bi2026motus}. Genie Envisioner adds the initial frame and sparsely sampled memory frames~\cite{liao2026genie}, and the RoboDojo recipe of VPP2 adds the initial frame and recent frames sampled at 1\,Hz~\cite{guo2026video}, close to our Window-history baseline. DreamZero uses autoregressive context and key-value caching~\cite{ye2026world}. Long-WAM extends the explicit history window and combines streaming visual encoding with visual key-value reuse within an action request~\cite{huang2026long}. The amount of context it processes depends on the selected window length.

Compressed alternatives include a Perceiver memory over buffered frame embeddings in MemoryVAM~\cite{jiang2026memoryvam}, recent, anchor, and gist tokens in a video-side cache in MemoryWAM~\cite{yang2026memorywam}, and novelty-merged event banks with progress supervision in DiM-WAM~\cite{wang2026dim}. Concurrently, RM-WAM injects event memory into the video branch of a Wan-based WAM through gated adapters~\cite{miao2026rm}. Recurrent states have also given video world models long-term memory~\cite{po2025long}.

A second question is which computation to reuse. Fast-WAM and mimic-video compute visual features once per action chunk and reuse them for action denoising~\cite{yuan2026fast,pai2025mimic}, while feature caching across denoising steps approximates network outputs that depend on the noise level~\cite{ma2024deepcache}. Concurrently, ActMem-VLA conditions only the early, high-noise denoising steps on an action-history memory~\cite{zhao2026remember}. In EWAM, video and action latents evolve jointly over denoising iterations~\cite{wang2026ewam}. We keep this joint process, maintain a memory state across requests, and reuse one current-conditioned readout within each request. Because the readout depends only on the strictly-past state and the current observation, reusing it at every iteration is exact. We evaluate the resulting performance-cost trade-off against No-history, Window-history, and Full-history inputs to the same backbone.

\section{Method}
\label{sec:method}

\subsection{Problem Formulation and Overview}
\label{sec:prelim}
\label{overview}
An action request must combine what the robot currently observes with evidence acquired earlier in the episode. Let $I_k$ be the current observation at the time $s_k$ of the $k$-th request, and let $\mathcal H_k=(I_{\tau_1},\ldots,I_{\tau_{n_k}})$ be the sampled history, with $\tau_1<\cdots<\tau_{n_k}<s_k$. For a nonempty history, $\tau_1=0$ denotes the initial frame of the episode. Given these observations and an instruction $\ell$, the policy predicts an action chunk. We seek a historical representation with three properties. Its state and the condition it supplies to generation do not grow with $n_k$, what it returns is determined by the current observation, and it depends only on observations strictly preceding $s_k$.

The memory module interacts with the generative backbone through two interfaces only. It takes patch tokens of observations and returns tokens that are appended to the video-side condition, so the backbone itself is unchanged. This applies to WAMs that generate future video from visual condition tokens and couple action generation to video generation. We instantiate it in EWAM~\cite{wang2026ewam}, which jointly denoises future-video latents and actions over several iterations per request, and whose Action Expert attends to the Video and Vision-Language Experts through asymmetric attention. Explicit history frames supplied as video-side conditions are processed by the backbone at every iteration, so their cost scales with both the history length and the number of denoising iterations. We address these costs with a compressed historical representation and incremental updates. The history available to a request stays fixed while the future-video and action latents are refined. We therefore incorporate newly sampled frames when a request begins, read the state once, and reuse that readout throughout denoising.

\begin{figure*}[t]
    \centering
    \includegraphics[width=\textwidth]{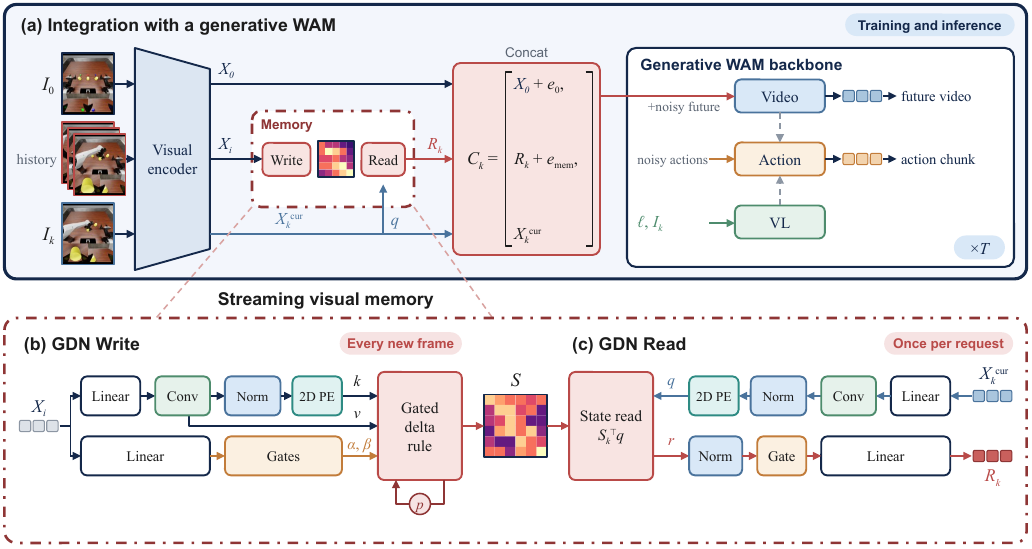}
    \caption{\textbf{Streaming visual memory.} (a) A shared visual encoder maps the initial frame $I_0$, the current observation $I_k$, and sampled history frames to patch tokens. History tokens $X_i$ are written into the memory, the current tokens $X_k^{\mathrm{cur}}$ query it, and the readout $R_k$ joins the initial frame and the current tokens in the condition $C_k$ of the Video Expert. The backbone is unchanged, and its Action Expert reads the Video and Vision-Language Experts through the existing attention. In our experiments the backbone is EWAM. (b) For each new frame, shared projections produce keys, values, and gates, and the gated delta rule decays the state once and then updates it patch by patch. (c) Once per request, each current patch forms a query from the same projections and its 2D position and reads the state, and a normalized, gated linear layer maps the readout to $R_k$. Boxes of the same color are modules of the same type. Thumbnails are Cover Blocks model-input canvases.}
    \label{fig:architecture}
\end{figure*}

Figure~\ref{fig:architecture} summarizes the resulting information flow. Sampled frames are encoded and written into a fixed-size memory state (\cref{compressed-visual-memory}). Patches of the current observation extract historical features from this state (\cref{current-conditioned-spatial-retrieval}). The memory readout, an initial frame, and the current observation are concatenated into the condition of the Video Expert (\cref{memory-conditioning-for-joint-videoaction-generation}). \Cref{streaming-memory-inference} describes how the state is maintained across requests and how the pathway is trained. Memory enters only through the video pathway. The Action Expert keeps its original structure and receives historical evidence through the existing asymmetric attention of EWAM.

\subsection{Compressed Visual Memory}
\label{compressed-visual-memory}
\label{visual-encoding-and-frame-wise-writing}
Retaining the entire sampled history as explicit frame tokens makes the generative context grow with the interaction. We instead represent sampled history by a matrix of key-value associations. Synchronized camera views are tiled into a single image, which a frozen VAE and a trainable patch projection encode as $X_i\in\mathbb R^{P\times d}$. In our RoboDojo setting, the head view and the two wrist views form a $384\times320$ canvas, which gives $P=120$ tokens of width $d=3072$. Retaining patch-level features allows the memory to associate local visual content with its image-plane location rather than reducing each observation to a single pooled vector.

We build on Gated DeltaNet (GDN)~\cite{yang2025gated}, which decays its state at every token, and instead separate retention across frames from updates within a frame. For clarity, we omit the memory-head index. Each observation $X$ is normalized and mapped to query, key, and value features,
\begin{equation}
[\widetilde Q,\widetilde K,V]=\operatorname{SiLU}\!\left(\operatorname{Conv}_4\!\left(\operatorname{RMSNorm}(X)\,W\right)\right),
\label{eq:projection}
\end{equation}
where $W$ is a bias-free linear map and $\operatorname{Conv}_4$ is a causal depthwise convolution of width four along the patch sequence, restarted at every frame. History frames supply keys and values for writing, and the current observation supplies queries through the same weights. Input-dependent gates produce a retention coefficient $\alpha_i\in(0,1)$ for each frame and a write strength $\beta_{i,p}\in(0,1)$ for each patch. Linear maps of the normalized patch features give gate logits $\hat\alpha_{i,p}$ and $\hat\beta_{i,p}$, and
\begin{equation}
\beta_{i,p}=\sigma(\hat\beta_{i,p}),\qquad
\alpha_i=\exp\!\Big(-\frac{e^{A_{\log}}}{P}\sum_{p=1}^{P}\operatorname{softplus}(\hat\alpha_{i,p}+c)\Big),
\label{eq:gates}
\end{equation}
with learned per-head scalars $A_{\log}$ and $c$. The frame retention averages the per-patch decay rates, so its strength does not depend on the patch count $P$. With state $S\in\mathbb R^{d_k\times d_v}$, key $k_{i,p}$, and value $v_{i,p}$, the update is
\begin{align}
S^{(i)}_0 &= \alpha_i S^{(i-1)}, \label{eq:frame_retention}\\
\delta v_{i,p} &= v_{i,p}-(S^{(i)}_{p-1})^{\top}k_{i,p}, \label{eq:write_residual}\\
S^{(i)}_p &= S^{(i)}_{p-1}+\beta_{i,p}k_{i,p}\delta v_{i,p}^{\top},
\label{eq:patch_write}
\end{align}
followed by $S^{(i)}=S^{(i)}_P$. The state starts at zero, and patches are processed in a fixed raster order.

The two operations serve different purposes. Frame-level retention scales the state carried into a frame once, independently of how many patches are then written. The delta term performs one step of online least-squares correction of the association $k\mapsto S^{\top}k$. The state predicts the value at the current key and writes the prediction residual, scaled by $\beta_{i,p}$~\cite{yang2025gated}. Associations that the state already predicts accurately produce small corrections. This residual form is useful for manipulation scenes with substantial visual redundancy across observations, motivating selective correction of the stored associations rather than repeated accumulation of every value. The matrix dimensions are independent of the number of history frames, and its contents form a learned summary. With 16 heads of size $d_k=d_v=128$, the FP32 state occupies 1\,MiB per environment. Appendix~B gives the multi-head implementation.

\subsection{Extracting History with the Current Observation}
\label{current-conditioned-spatial-retrieval}
Different decisions require different parts of the same history. In Cover Blocks, for example, which cup to lift next depends on the color of the block that was visible at that location before it was covered. We therefore let each patch of the current observation extract a patch-aligned historical feature. Let $\widetilde q_{k,p}$ be a query computed from current patch $p$, and let $\rho_p$ denote the L2-normalized two-dimensional sinusoidal encoding of its position on the $12\times10$ patch grid. Keys, queries, and the readout are
\begin{align}
k_{i,p} &= \mathcal N_2\!\left(\mathcal N_2(\widetilde k_{i,p})+\eta\rho_p\right),\\
q_{k,p} &= \mathcal N_2\!\left(\mathcal N_2(\widetilde q_{k,p})+\eta\rho_p\right),\\
r_{k,p} &= S_k^{\top}q_{k,p}/\sqrt{d_k},
\label{eq:memory_read}
\end{align}
where $\mathcal N_2$ denotes unit Euclidean normalization, $\eta$ is a learned per-head scale initialized to 0.1, and $S_k=S^{(n_k)}$ summarizes only observations preceding $s_k$. Historical keys use the same positional construction, so an association is addressed by both visual appearance and image-plane location, whereas GDN keys and queries carry no positional term. The content term lets a query match objects by appearance. The positional term biases it toward what was previously observed at the same image location, which corresponds to the same workspace region when the viewpoint changes little.

Readouts from the memory heads are normalized, gated by the current features, and projected into the video-token space, yielding $R_k\in\mathbb R^{P\times d}$ with one historical feature per current patch. With $h$ indexing the heads,
\begin{equation}
R_{k,p}=\big[\operatorname{RMSNorm}(r^{(h)}_{k,p})\odot\operatorname{SiLU}(z^{(h)}_{k,p})\big]_{h=1}^{16}\,W_o,
\label{eq:readout_projection}
\end{equation}
where the output gate $z_{k,p}$ is a linear map of the normalized current patch features, $[\cdot]_h$ concatenates the heads into 2048 features, and $W_o\in\mathbb R^{2048\times d}$ is initialized to zero, so the memory tokens carry no content at the start of training. In GDN, each token queries the state right after writing itself. Here the current observation supplies the query without first being written into the state, so the readout contains only strictly past evidence. If the current frame lies on the history-sampling grid, it is written at a later request.

\subsection{Integrating Memory into Joint Generation}
\label{memory-conditioning-for-joint-videoaction-generation}
The memory readout, the initial frame, and the current observation play different roles in a decision. We supply them to the Video Expert as separate groups of tokens,
\begin{equation}
C_k=\left[X_0+e_0,\ R_k+e_{\mathrm{mem}},\ X_k^{\mathrm{cur}}\right],
\label{eq:three_block_condition}
\end{equation}
where $[\cdot]$ concatenates along the token dimension, $X_0$ contains the features of the episode's initial frame $I_0$, $X_k^{\mathrm{cur}}$ contains the current visual features, and learned type embeddings $e_0$ and $e_{\mathrm{mem}}$ distinguish the initial-frame tokens and memory tokens. The current tokens are the unchanged current-frame tokens of EWAM and provide the evidence needed for the immediate action. The memory tokens summarize the sampled history. The initial frame preserves direct access to the starting scene, where several of our tasks present information needed later, such as the colors and arrangement of objects. Because the memory state decays and has limited capacity, the initial-frame tokens keep the starting scene available at a fixed cost of $P$ tokens. The initial frame is also written into memory once it becomes strictly past. Appending memory as separate tokens, rather than adding it to the current tokens, leaves the current features intact and lets attention in the Video Expert determine how much historical context each layer uses.

At each denoising iteration, the Video Expert processes $C_k$ together with the evolving future-video latents, and the Action Expert attends to the resulting video-side representations alongside the vision-language and action features of EWAM. The auxiliary tokens act only as context. They are excluded from video prediction targets and latent updates, and their keys are masked when no valid history exists at the start of an episode. The initial-frame and memory tokens each use a single-frame rotary position grid, while the current and future frames keep the original grid of EWAM. With the current frame and two future latent frames, the video sequence has $5P=600$ tokens. Appendix~A compares this concatenation with residual addition of the memory to the current tokens. Appendix~C gives the masks and positional treatment.

\begin{table*}[t]
\centering
\small
\setlength{\tabcolsep}{5pt}
\caption{Closed-loop success on six RoboDojo tasks with checkpoints trained for \mainbudget{} updates. Each task contains 50 fixed evaluation layouts, and Overall is the success rate over all 300 episodes. All methods use the same backbone initialization, training split, global batch size, paired evaluation layouts, and paired action noise.}
\label{tab:main_results}
\begin{tabular}{@{}lccccccc@{}}
\toprule
Method & \shortstack{Cover\\Blocks} & \shortstack{Match \& Pick\\from Conveyor} & \shortstack{Press by\\Number} & \shortstack{Imitate Sorting\\Sequence} & \shortstack{Swap\\Blocks} & Swap T & Overall (\%) \\
\midrule
No-history & 12 & 11 & 0 & 0 & 0 & 0 & 7.67 \\
Window-history & 39 & 12 & 0 & 0 & 3 & 1 & 18.33 \\
Full-history & 35 & 11 & 2 & 0 & 2 & 0 & 16.67 \\
Streaming-history (ours) & 49 & 11 & 0 & 0 & 3 & 1 & 21.33 \\
\bottomrule
\end{tabular}
\end{table*}

\subsection{Efficient Streaming Inference and Learning}
\label{streaming-memory-inference}
Streaming inference combines incremental updates across requests with readout reuse within a request. Let $\Delta n_k=n_k-n_{k-1}$ be the number of newly sampled history frames, with $n_0=0$ and $S_0=0$. At the start of a request, the state advances only through these frames, in chronological order,
\begin{equation}
S_k=\mathcal U_\theta\!\left(S_{k-1},
X_{n_{k-1}+1},\ldots,X_{n_k}\right),
\label{eq:stream_update}
\end{equation}
where $\mathcal U_\theta$ applies Eqs.~\ref{eq:frame_retention} to \ref{eq:patch_write}. If no new frame has been sampled, the state is retained. The current observation then produces one readout, from which $C_k$ is formed and reused at every denoising iteration. With $T$ iterations, the request computes
\begin{equation}
(y_{j-1},a_{j-1})=\Phi_\theta\!\left(y_j,a_j,j,C_k,\ell\right)
\label{eq:denoise}
\end{equation}
for $j=T,\ldots,1$, where $y_j$ and $a_j$ are the noisy future-video latents and actions and $\Phi_\theta$ is one joint iteration of the backbone. The backbone still recomputes its layerwise representations from this condition and the evolving noisy latents. The state is retained until the episode resets. Writes run synchronously at the start of each request, and a reset clears the state, the unwritten history, and the initial-frame cache.

The compressed condition and the incremental updates control different sources of cost. Across $K$ requests, incremental updates perform $\sum_{k=1}^K\Delta n_k=n_K$ frame writes, whereas reconstructing the state from the full prefix at every request performs $\sum_{k=1}^K n_k$ writes. Within a request, the memory is read once. The historical context processed at every denoising iteration consists of the $2P$ auxiliary tokens of $C_k$ at a fixed observation resolution, whereas explicit full history contributes $n_kP$ tokens per iteration. The rotary position embedding (RoPE) tables are also fixed within a request and can be cached across layers and denoising iterations, which we refer to as RoPE caching. \Cref{streaming-inference-efficiency} measures the resulting latency and memory, and Appendix~D gives the sampling and state-management protocol.

During training, each sampled example reconstructs its memory from the associated historical prefix. Gradients of the joint video and action flow-matching objectives~\cite{lipman2022flow} pass through the conditioning interface into memory reads and writes, so the history representation is learned for the prediction objectives of the backbone without memory-specific labels. The two objectives have equal weight, $\mathcal L=\mathcal L_{\mathrm{video}}+\mathcal L_{\mathrm{action}}$. At inference, the same recurrence is maintained incrementally across requests. Appendix~E specifies the backbone assets, trainable components, and optimization settings.

\section{Experiments}
\label{sec:experiments}

\begin{figure*}[t]
\centering
\includegraphics[width=\textwidth]{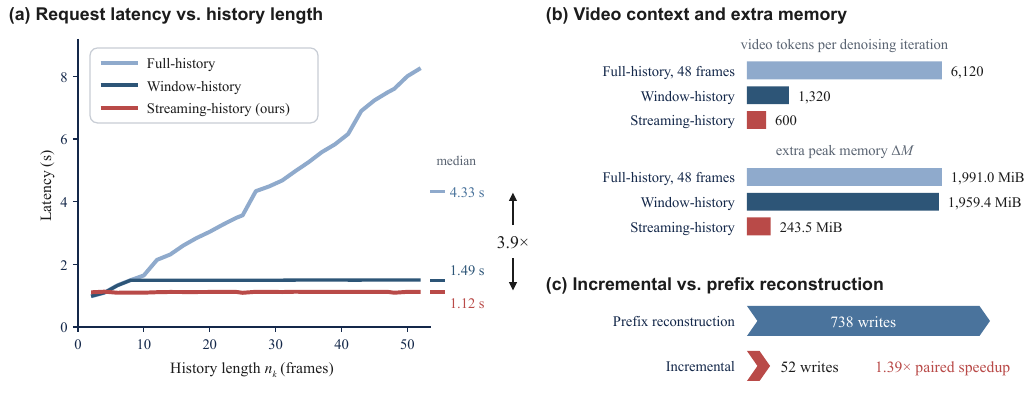}
\caption{\textbf{Three aspects of inference efficiency.} (a) Request latency against history length on a common trajectory, with RoPE caching for all methods. Curves are per-index medians over three measurement repeats per method. Median labels give the median request latency of each method, and the arrow gives the $3.9\times$ paired per-request speedup over Full-history. (b) Video tokens per denoising iteration and the peak-allocation increment $\Delta M$ from a length sweep with one fixed model, relative to the 360-token floor. Full-history uses $n_k=48$, and Window-history uses a full window. (c) Frame writes per trajectory for incremental updates and prefix reconstruction with the same weights on three trajectories, both with RoPE caching, together with the median paired speedup. All measurements use one RTX 5090.}
\label{fig:quality_cost}
\end{figure*}

We evaluate whether compressed history supports closed-loop manipulation and how streaming inference changes its cost. The four-way comparison measures the performance-cost trade-off among history representations, and a same-checkpoint control isolates the effect of incremental updates.

\subsection{Experimental Setup}
\label{experimental-setup-and-baselines}

\paragraph{Benchmark.}
We use six memory-dependent RoboDojo tasks, namely Cover Blocks, Match and Pick from Conveyor, Press by Number, Imitate Sorting Sequence, Swap Blocks, and Swap T. They require recalling object identity or location, demonstrated order, or progress through a manipulation sequence. Each task uses 50 fixed evaluation layouts, giving 300 episodes per method. We report the binary episode success returned by the simulator and average it over all six tasks. Task-specific success conditions and episode limits are given in Appendix~F. The policy predicts 16 action targets at three-frame intervals, which are interpolated to 48 control steps at 25\,Hz.

\paragraph{Compared methods.}
All methods share the EWAM backbone, built on the Wan2.2-TI2V-5B video model and the Qwen3-VL-2B vision-language model~\cite{team2025wan,bai2025qwen3}, and differ in their history conditioning. \textbf{No-history} uses only the current observation. \textbf{Window-history} supplies the initial frame and up to seven recent observations sampled at 1\,Hz, deduplicating overlapping frames. \textbf{Full-history} supplies all observations sampled at 1\,Hz before the current observation. \textbf{Streaming-history} uses our condition $C_k$, which concatenates an initial frame, a memory readout, and the current observation (\cref{memory-conditioning-for-joint-videoaction-generation}). No-history is the baseline without history. Window-history tests compressed memory against a bounded explicit context, and Full-history compares it with access to the complete sampled history.

\paragraph{Training and evaluation.}
For closed-loop evaluation, we use checkpoints trained on A800 GPUs for \mainbudget{} optimizer updates. All configurations fine-tune the same EWAM RoboDojo checkpoint on a shared episode-level split of 90 training and 10 held-out demonstrations per task, with global batch size 64 and training seed 20261024. Evaluation uses the same layouts and paired action noise across methods. Appendix~E provides the training settings, and Appendix~F details the evaluation protocol.

\subsection{Closed-Loop Performance}
\label{closed-loop-performance}
\Cref{tab:main_results} compares all six tasks at the same training budget. Streaming-history achieves the highest overall success, completing 64 of 300 episodes (21.33\%). No-history, Window-history, and Full-history complete 23 (7.67\%), 55 (18.33\%), and 50 (16.67\%) episodes, respectively. Streaming-history improves on No-history by 13.67 percentage points and on Window-history by 3.00 points. It also exceeds Full-history by 4.67 points while using a fixed-length video sequence.

Cover Blocks provides the clearest comparison. Streaming-history succeeds in 49 of 50 layouts, compared with 12 for No-history, 39 for Window-history, and 35 for Full-history. Window-history already retains the initial frame, so this comparison evaluates the learned memory representation against the initial frame plus recent explicit frames. Together, these results support compressed history as an effective alternative to extending the explicit visual context.

Appendix~A compares our concatenated condition with a separately trained variant that adds the memory readout to the current tokens. This residual addition succeeds in 54 of 300 episodes, compared with 64 for concatenation. Appendix~H reports experiments on three tabletop manipulation tasks and three variants of the first task on a Unitree G1-D robot.

\subsection{Inference Cost}
\label{streaming-inference-efficiency}
We evaluate three aspects of inference efficiency, namely request latency across history representations, the memory cost of the denoising context, and the cost of maintaining one memory model incrementally.

\paragraph{Serving performance.}
We measure complete action requests on one RTX 5090 using a common recorded trajectory, BF16 inference, and ten denoising iterations. All methods use RoPE caching, and Appendix~G specifies the measurement boundary. We report the median of three per-run statistics for every method.

\Cref{fig:quality_cost}a shows a median request latency of 1.119\,s for Streaming-history, compared with 1.492\,s for Window-history and 4.333\,s for Full-history. Pairing requests by index gives a 25\% latency reduction relative to Window-history and a $3.9\times$ speedup over Full-history. The complete 28-request trajectory requires 31.1\,s of inference, versus 40.2\,s and 117.5\,s, respectively. These costs complement the closed-loop comparison, since compressed history improves overall success while reducing the context processed at each request.

\begin{table}[t]
\centering
\small
\setlength{\tabcolsep}{4pt}
\caption{Denoising context and additional peak allocation. $L$ is the video sequence length. $\Delta M$ is measured using one Full-history model at the corresponding length, relative to its 360-token floor, and includes historical encoding and temporary workspaces. The Full-history point uses $n_k=48$, and Window-history uses a full window.}
\label{tab:context_memory}
\begin{tabular}{@{}lcr@{}}
\toprule
Method & Video tokens $L$ & $\Delta M$ (MiB) \\
\midrule
Window-history & $1320$ & 1959.4 \\
Full-history & $360+120n_k$ & 1991.0 \\
Streaming-history & $600$ & \textbf{243.5} \\
\bottomrule
\end{tabular}
\end{table}

\paragraph{Scaling with history.}
For histories of 1 to 10, 11 to 30, and 31 to 52 frames, Streaming-history has median latencies of 1.096, 1.117, and 1.121\,s, whereas Full-history grows from 1.327 to 3.150 and 6.524\,s (\cref{fig:quality_cost}a). The long-history interval yields a $5.8\times$ paired per-request speedup. Window-history reaches a plateau once its eight-frame context fills.

\paragraph{Latency breakdown.}
Joint denoising takes 981.1, 1317.0, and 3984.7\,ms of a profiled Streaming-history, Window-history, and Full-history request, which is 89\%, 90\%, and 92\% of the respective totals. Memory maintenance adds 33.5\,ms of state writes and 1.1\,ms of readout per request. Historical encoding takes 22.6\,ms, compared with 69.9\,ms for the window and 254.1\,ms for full history. The latency gap therefore arises mainly from the context processed in every denoising iteration, which the compressed condition keeps at a fixed length.

\paragraph{Context length and memory.}
Streaming-history fixes the video sequence at 600 tokens, compared with 1,320 for a full Window-history context and $360+120n_k$ for Full-history (Table~\ref{tab:context_memory}). We isolate the additional allocation associated with these lengths using one fixed model and subtract its no-history, 360-token peak. The resulting increment is 243.5\,MiB at 600 tokens, versus 1,959.4\,MiB at 1,320 and 1,991.0\,MiB at 6,120 (\cref{fig:quality_cost}b). The compact context therefore reduces the measured additional allocation by approximately 1.7\,GiB. Beyond eight explicit history frames, the measured peak largely plateaus under chunked encoding and fused attention, while request latency continues to grow. Appendix~G provides the full sweep and allocation boundary.

\paragraph{Incremental updates.}
To isolate state maintenance, we compare incremental updates with prefix reconstruction using the same Streaming-history checkpoint, recorded inputs, and shared cache of historical latents. Both modes use RoPE caching. Across three trajectories, incremental updates perform 52 frame writes per trajectory rather than 738, with a $1.39\times$ median speedup over 81 paired non-initial requests (\cref{fig:quality_cost}c). All 84 paired action hashes and the 81 non-initial state hashes agree, confirming that the measured gain comes from eliminating repeated state construction.

\section{Limitations and Future Work}
\label{sec:limitations}

The fixed-capacity memory trades direct access to individual past frames for a compact representation. Observations are sampled at 1\,Hz, which can miss brief events between sampling points. Adaptive sampling could capture such events more effectively. The memory read associates content with two-dimensional image-plane positions, so large viewpoint changes remain a challenge. More expressive recurrent updates~\cite{siems2026deltaproduct} and object-level correspondence are possible extensions.

Counting and reproducing long demonstrated sequences remain challenging at the evaluated 30K checkpoint. Auxiliary memory objectives or explicit progress representations~\cite{chen2026rmbench,zhang2026mem4act} offer directions for improving temporal reasoning.

New history is written synchronously at the start of a request, so longer gaps increase the next request's preprocessing work. Asynchronous execution such as real-time chunking~\cite{black2025realtimeexecutionactionchunking} could overlap these writes with action execution. Joint video-action denoising still processes the evolving future latents at every iteration, making denoising acceleration complementary to incremental memory updates. The memory module requires only patch tokens and a video-side condition, and our evaluation covers one generative WAM, six simulation tasks, and three tabletop tasks plus three variants of the first task on a single robot platform (Appendix~H). Applying it to other WAM backbones would further test this transferability.

\section{Conclusion}
\label{sec:conclusion}

We presented streaming visual memory for generative world-action models, a compressed memory module that attaches to the video-side condition without modifying the backbone, together with an efficient procedure for streaming inference. A fixed-size memory state accumulates sampled visual history, and a current-conditioned readout supplies a fixed number of memory tokens to joint video-action generation alongside an initial frame. Incremental updates and one readout per request reduce repeated history processing while keeping the denoising context fixed in length. On six RoboDojo tasks at the same update budget and global batch size, Streaming-history achieves 21.3\% success, compared with 7.7\% for No-history, 18.3\% for Window-history, and 16.7\% for Full-history. Serving measurements on one RTX 5090 with RoPE caching for every method give a median request latency of 1.119\,s, 25\% below Window-history and $3.9\times$ faster than Full-history.

{\small
\bibliographystyle{ieeenat_fullname}
\bibliography{main}
}
\onecolumn
\section*{Appendix}
\appendix
\numberwithin{table}{section}
\numberwithin{figure}{section}

This supplement is organized as follows. Appendix~\ref{app:interfaces} compares concatenation with residual addition. Appendices~\ref{app:gdn} to~\ref{app:streaming} detail the memory formulation, visual encoding, and streaming protocol. Appendix~\ref{app:training} lists the training configuration, Appendix~\ref{app:results} the simulation protocol and paired outcomes, and Appendix~\ref{app:profiling} the latency and memory measurements. Appendix~\ref{app:robot} reports the real-robot experiments.

\section{Concatenation versus Residual Addition}
\label{app:interfaces}
Streaming-history supplies an initial frame, a memory readout, and the current observation as separate token groups. We compare this concatenation with residual addition, which adds projected memory features directly to the current-observation tokens without separate initial-frame or memory tokens (Table~\ref{tab:interface_identity}).

\begin{table*}[htbp]
\centering
\small
\caption{Two ways of supplying the memory. Token counts include the current frame, two future latent frames, and the auxiliary tokens, with 120 tokens per frame.}
\label{tab:interface_identity}
\begin{tabular}{@{}l l r@{}}
\toprule
Variant & Video-side sequence & Tokens \\
\midrule
Concatenation (ours) & Initial frame, memory, current observation, future latents & 600 \\
Residual addition & Current observation plus memory residual, future latents & 360 \\
\bottomrule
\end{tabular}

\vspace{1.2em}
\centering
\small
\caption{Concatenation versus residual addition with checkpoints trained for 30K updates. Entries give successes out of 50 per task and out of 300 overall.}
\label{tab:interface_success}
\begin{tabular}{@{}lccccccc@{}}
\toprule
Variant & \shortstack{Cover\\Blocks} & \shortstack{Match and Pick\\from Conveyor} & \shortstack{Press by\\Number} & \shortstack{Imitate Sorting\\Sequence} & \shortstack{Swap\\Blocks} & Swap T & Overall \\
\midrule
Concatenation (ours) & 49 & 11 & 0 & 0 & 3 & 1 & 64 \\
Residual addition & 40 & 13 & 1 & 0 & 0 & 0 & 54 \\
\bottomrule
\end{tabular}
\end{table*}

\paragraph{Separate tokens versus residual addition.}
Under matched data splits, seeds, optimizer settings, and a 30K-update budget (protocol in Appendix~\ref{app:results}), concatenation reaches 21.33\% overall success compared with 18.00\% for residual addition (Table~\ref{tab:interface_success}). Concatenation leads on Cover Blocks, Swap Blocks, and Swap T, whereas residual addition leads on Match and Pick from Conveyor and Press by Number.

\section{Detailed Memory Formulation}
\label{app:gdn}
\begin{figure*}[htbp]
\centering
\includegraphics[width=\textwidth]{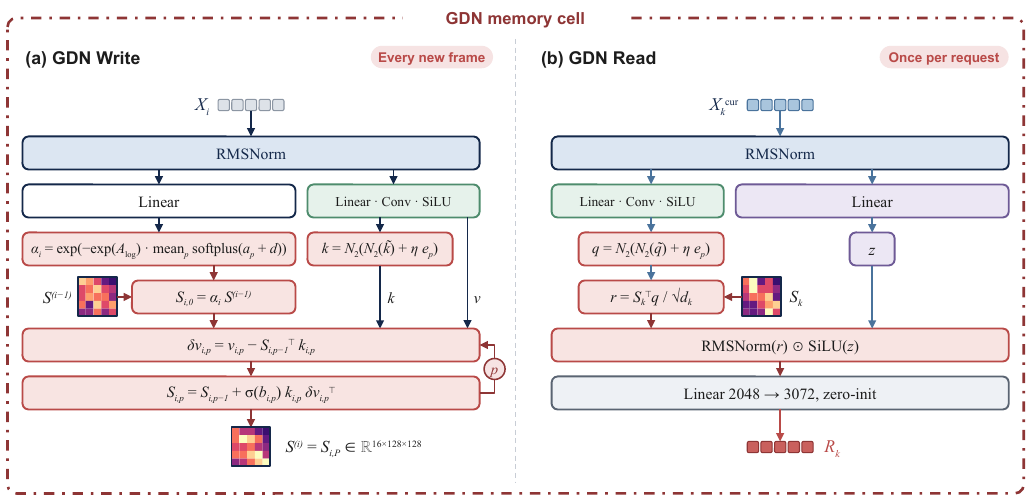}
\caption{\textbf{GDN memory cell.} (a) Each history frame is normalized and projected. The key combines the content feature with the 2D position, the retention coefficient $\alpha_i$ is computed once per frame, and patches then update the state in raster order with the delta rule. (b) The current observation reuses the normalization and projections of (a), reads the strictly-past state $S_k$ once per request, and maps the gated readout to $R_k$ with a zero-initialized projection.}
\label{fig:supp_gdn}
\end{figure*}

We build on Gated DeltaNet (GDN)~\cite{yang2025gated}. Figure~\ref{fig:supp_gdn} summarizes the Streaming-history memory cell.

For each history frame, let \(X_i\in\mathbb{R}^{P\times3072}\) contain the visual patch tokens and let \(\bar X_i=\operatorname{RMSNorm}(X_i)\). A bias-free linear map produces concatenated query, key, and value features. A causal depthwise convolution of width four, followed by SiLU, acts along the frame's patch sequence and restarts at each frame without a cross-frame cache. The result splits into \(\widetilde Q_i,\widetilde K_i,V_i\), each with 16 heads of dimension 128. Separate linear maps of \(\bar X_i\) produce gate logits \(\hat\alpha_{i,p}^{(h)},\hat\beta_{i,p}^{(h)}\) and the output gate \(z_{i,p}^{(h)}\).

Let \(\rho_p\) be the L2-normalized two-dimensional sinusoidal position encoding on the patch raster, and let \(\mathcal N_2\) denote L2 normalization. Keys and queries use
\[
k_{i,p}^{(h)}=\mathcal N_2(\mathcal N_2(\widetilde k_{i,p}^{(h)})+\eta_h\rho_p),
\]
with the same construction for \(q\) and a learned per-head positional scale \(\eta_h\) initialized to 0.1. With learned parameters \(A_{\log}^{(h)}\) and \(c^{(h)}\), the gates are
\[
\beta_{i,p}^{(h)}=\sigma(\hat\beta_{i,p}^{(h)}),
\]
\[
\alpha_i^{(h)}=\exp\!\left[-\frac{\exp(A_{\log}^{(h)})}{P}
\sum_{p=1}^P\operatorname{softplus}(\hat\alpha_{i,p}^{(h)}+c^{(h)})\right].
\]
Averaging the transformed decay logits defines one retention factor per frame, separating explicit frame-level decay from the subsequent patch updates. The incoming state is decayed once, \(S^{(i,h)}_0=\alpha_i^{(h)}S^{(i-1,h)}\), and patches then update it in raster order,
\[
\delta v_{i,p}^{(h)}=v_{i,p}^{(h)}-(S^{(i,h)}_{p-1})^\top k_{i,p}^{(h)},
\]
\[
S^{(i,h)}_p=S^{(i,h)}_{p-1}+\beta_{i,p}^{(h)}k_{i,p}^{(h)}(\delta v_{i,p}^{(h)})^\top.
\]
The final state is \(S^{(i,h)}=S^{(i,h)}_P\). Invalid padded frames neither decay nor update the state, which starts at zero in FP32. Because patch updates run sequentially, the result depends on patch order.

The current query reads the strictly-past memory state constructed only from earlier observations,
\[
r_{k,p}^{(h)}=(S_k^{(h)})^\top q_{k,p}^{(h)}/\sqrt{128}.
\]
Each head's readout undergoes RMS normalization with a learned 128-dimensional scale shared across heads, followed by elementwise multiplication with \(\operatorname{SiLU}(z_{k,p}^{(h)})\). Concatenating heads yields 2048 features per patch. Streaming-history applies a zero-initialized \(2048\!\to\!3072\) memory-content projection so that memory tokens carry no content at initialization, whereas residual addition uses its own output projection to form a residual on the current observation. We implement the recurrence with the flash-linear-attention library~\cite{yang2024fla}.

\section{Visual Encoding and Implementation Details}
\label{app:visual}

\begin{table}[htbp]
\centering
\small
\caption{RoboDojo implementation settings for Streaming-history.}
\label{tab:implementation_settings}
\begin{tabular}{@{}p{0.34\columnwidth}p{0.60\columnwidth}@{}}
\toprule
Parameter & Setting \\
\midrule
Video assets & Wan2.2-TI2V-5B~\cite{team2025wan} \\
Vision-language assets & Qwen3-VL-2B-Instruct~\cite{bai2025qwen3} \\
Action width / layers & 1024 / 30 \\
RGB input ($H\times W$) & $384\times320$ \\
VAE latent grid & $24\times20$ \\
Patch grid / count & $12\times10$ / $P=120$ \\
Video-token width & $d=3072$ \\
Memory heads & 16 \\
Key / value width & $d_k=d_v=128$ \\
Patch convolution & Width 4, reset per frame \\
Initial positional scale & $\eta_h=0.1$ \\
Memory projection & $2048\to3072$, zero initialized \\
State precision / size & FP32 / 1 MiB per environment \\
History sampling & 1 Hz, strictly past \\
RoboDojo observation rate & 25 Hz \\
\bottomrule
\end{tabular}
\end{table}

Table~\ref{tab:implementation_settings} lists the RoboDojo implementation settings. Synchronized camera views are tiled into a single image by stacking a 240-by-320 head view above two 120-by-160 wrist views (left, then right) with twelve-pixel top and bottom padding. Streaming-history encodes the initial frame and each history frame independently through the frozen visual VAE, processing one frame per call and sharing a trainable patch projection across observations.

Streaming-history supplies \([X_0+e_0,R_k+e_{\mathrm{mem}},X_k^{\mathrm{cur}},X_{\mathrm{future}}]\), where brackets denote token-wise concatenation. Learned, zero-initialized type embeddings $e_0$ and $e_{\mathrm{mem}}$ distinguish the initial-frame and memory tokens while leaving the current and future tokens unchanged. The episode's initial frame is encoded once, reused across requests, and written into memory once it becomes strictly past.

\begin{figure*}[htbp]
\centering
\includegraphics[width=\textwidth]{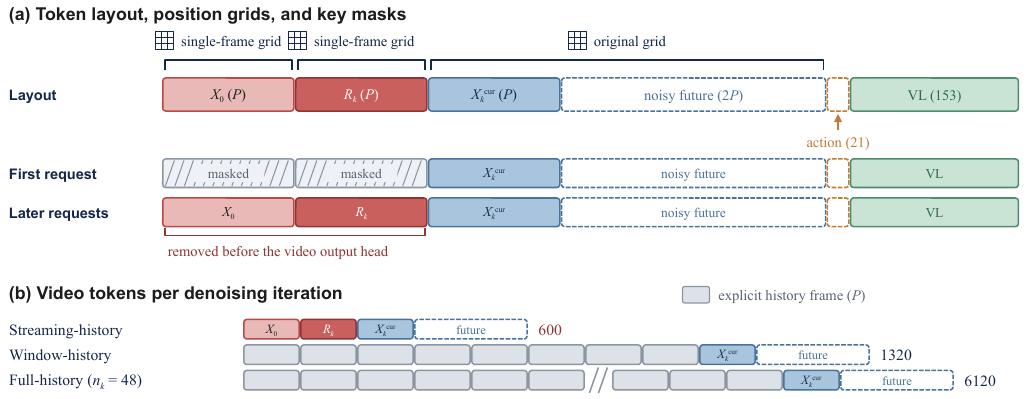}
\caption{\textbf{Video-side token layout and masks.} (a) The condition $C_k$ places the initial frame and the memory readout, each on a single-frame rotary grid, before the current and noisy future latents, which keep the original grid. Their keys are masked at the first request, and both are removed before the video output head. (b) Video tokens per denoising iteration. Window-history holds up to eight explicit frames, and Full-history grows by $P$ tokens per sampled frame.}
\label{fig:supp_tokens}
\end{figure*}

Memory validity is governed by history length during training and by completed writes during serving, rather than by state values. At episode start, no valid history exists, and the initial-frame tokens become valid only after the first request. When auxiliary tokens are invalid, their keys are masked from all queries, while their own queries still execute and their outputs are discarded. The initial-frame and memory tokens each use a single-frame rotary position grid, whereas the current and future frames preserve their original grid. Auxiliary hidden features and their time embeddings are stripped before the video output head, excluding them from video prediction targets and latent updates (Figure~\ref{fig:supp_tokens}a).

The Wan VAE encodes the current frame into one latent and compresses the eight predicted RGB frames into two future latents, one per four frames. The 16 action targets at stride 3 are interpolated at execution into the 48 control steps returned to the controller. Combining the 360 base tokens from the current and two future latent frames with 240 auxiliary tokens (120 initial-frame and 120 memory) yields 600 video tokens. Adding 21 action tokens and 153 vision-language tokens produces a joint sequence of 774 tokens, compared with 1494 for a full Window-history context.

Residual addition shares this camera layout, latent resolution, and trainable patch projection. It projects a single current-conditioned readout to $d=3072$ and adds it to the current-observation tokens at each denoising iteration without accumulation, keeping the video sequence at 360 tokens.

\section{Streaming Inference Protocol}
\label{app:streaming}
\begin{figure*}[htbp]
\centering
\includegraphics[width=\textwidth]{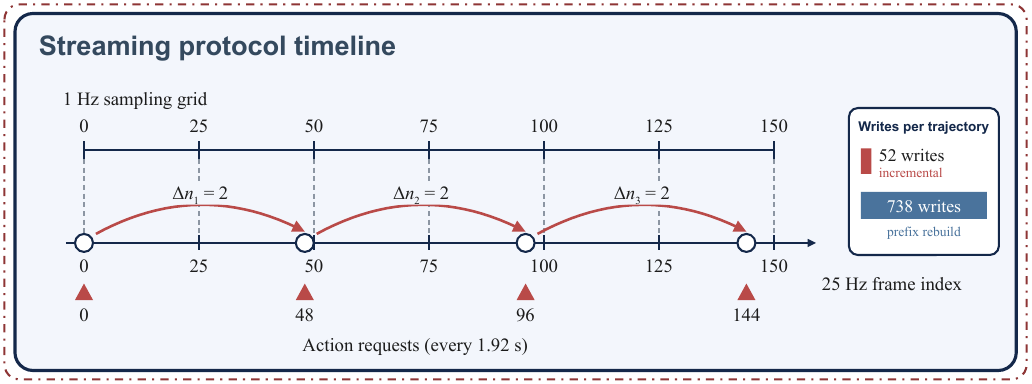}
\caption{\textbf{Streaming protocol.} History is sampled on the 1 Hz grid of the 25 Hz frame index, and actions are requested every 48 frames. Each request writes only the $\Delta n$ frames sampled since the previous request, so a 28-request trajectory performs 52 writes, whereas rebuilding the state from the full prefix at every request performs 738.}
\label{fig:supp_timeline}
\end{figure*}

For 25 Hz observations, training samples use the 1 Hz native-frame grid \(0,25,50,\ldots<\mathrm{anchor}\). Online sampling follows elapsed episode time from zero with strictly increasing timestamps. The current observation is excluded from the memory used by its own request, even at a sampling boundary. Figure~\ref{fig:supp_timeline} illustrates the first 150 frames of a trajectory.

During an episode, each action request executes four steps (Algorithm~\ref{alg:request}).
\begin{enumerate}
\tightlist
\item Encode newly sampled history frames and update the memory state in temporal order.
\item Retain the resulting state, clear the buffer of unwritten frames, and increment the completed-write count.
\item Encode the current observation and construct one readout and the auxiliary tokens.
\item Reuse these tokens at every denoising iteration while the Video and Action Experts recompute layerwise hidden states and attention.
\end{enumerate}

\begin{algorithm}[htbp]
\caption{One request of Streaming-history, written for a single memory head. Per-head readouts are concatenated before $W_o$.}
\label{alg:request}
\small
\begin{algorithmic}[1]
\For{each newly sampled frame $X_i$, in temporal order}
\State $[\widetilde Q,\widetilde K,V]\gets\operatorname{SiLU}(\operatorname{Conv}_4(\operatorname{RMSNorm}(X_i)\,W))$
\State $\beta_{i,p}\gets\sigma(\hat\beta_{i,p})$ and compute $\alpha_i$ from the averaged decay logits
\State $k_{i,p}\gets\mathcal N_2(\mathcal N_2(\widetilde k_{i,p})+\eta\rho_p)$ for every patch $p$
\State $S\gets\alpha_i S$ \Comment{decay once per frame}
\For{$p=1,\ldots,P$ in raster order}
\State $\delta\gets v_{i,p}-S^{\top}k_{i,p}$
\State $S\gets S+\beta_{i,p}\,k_{i,p}\,\delta^{\top}$
\EndFor
\EndFor
\State $q_{k,p}\gets\mathcal N_2(\mathcal N_2(\widetilde q_{k,p})+\eta\rho_p)$ from the current tokens
\State $r_{k,p}\gets S^{\top}q_{k,p}/\sqrt{d_k}$ \Comment{one read per request}
\State $R_{k,p}\gets[\operatorname{RMSNorm}(r^{(h)}_{k,p})\odot\operatorname{SiLU}(z^{(h)}_{k,p})]_{h}\,W_o$ \Comment{per-head readouts are concatenated before $W_o$}
\State $C_k\gets[X_0+e_0,\ R_k+e_{\mathrm{mem}},\ X_k^{\mathrm{cur}}]$
\State draw the initial future-video latents $y_T$ and actions $a_T$ from Gaussian noise
\For{$j=T,\ldots,1$}
\State $(y_{j-1},a_{j-1})\gets\Phi_\theta(y_j,a_j,j,C_k,\ell)$ \Comment{$C_k$ is reused}
\EndFor
\State return the actions $a_0$ and keep $S$ for the next request
\end{algorithmic}
\end{algorithm}

Writes execute synchronously at request start. Reset clears the memory state, unwritten history, write count, and initial-frame cache. As a reference for incremental updates, prefix reconstruction rebuilds the state from the entire prefix at each request using the same checkpoint and inputs (Appendix~\ref{app:profiling}).

\section{Training Configuration}
\label{app:training}
All closed-loop checkpoints are trained on A800 GPUs from the shared EWAM RoboDojo checkpoint (backbone assets in Table~\ref{tab:implementation_settings}). The four main methods and the residual-addition variant share the episode-level data split, seed, global batch size, and optimization schedule listed in Table~\ref{tab:training_settings}.

\begin{table}[htbp]
\centering
\small
\caption{Training and evaluation settings. The reported checkpoints are selected at 30K optimizer updates.}
\label{tab:training_settings}
\begin{tabular}{@{}p{0.35\columnwidth}p{0.59\columnwidth}@{}}
\toprule
Parameter & Setting \\
\midrule
Demonstrations (train / held out) & 90 / 10 per task, 540 / 60 across six tasks \\
Sampling & Uniform over demonstration episodes, strictly-past 1 Hz history \\
Prediction targets & 16 actions at stride 3, 8 video frames, one per two action targets \\
Initialization & Shared EWAM RoboDojo checkpoint \\
Random seed & 20261024 \\
Global batch & 64 \\
No-history batch & 8 ranks $\times$ 8 samples $\times$ 1 accumulation step \\
Window-history / Streaming-history / residual-addition batch & 16 ranks $\times$ 4 samples $\times$ 1 accumulation step \\
Full-history batch & 32 ranks $\times$ 2 samples $\times$ 1 accumulation step \\
Optimizer & AdamW, learning rate $5\times10^{-5}$, betas $(0.9,0.95)$ \\
Weight decay & 0.01, with exemptions described below \\
Learning-rate schedule & Linear warmup over 200 optimizer updates \\
Loss weights (video / action) & 1 / 1 \\
Precision & BF16, FP32 memory state \\
Memory / optimizer handling & History-frame activation checkpointing for the memory state, ZeRO-1 \\
Gradient clipping & 0.5 \\
Evaluated checkpoint & 30000 optimizer updates \\
\bottomrule
\end{tabular}
\end{table}

For Streaming-history, the VAE and Qwen visual encoder remain frozen. We train the video backbone and patch projection, the Qwen language backbone and mixture-of-transformers projections, the Action Expert, the memory, the memory-content projection, and the applicable type embeddings. Memory retention parameters, normalization weights, and positional scales are exempt from weight decay.

\section{Simulation Protocol and Extended Results}
\label{app:results}
We evaluate on the Memory dimension of RoboDojo~\cite{chen2026robodojo}, which draws on RMBench~\cite{chen2026rmbench}, through the XPolicyLab client-server interface~\cite{chen2026xpolicylab}. Each task uses 50 fixed evaluation layouts, each contributing one scored episode (Table~\ref{tab:sim_protocol}). The simulator's reward manager verifies completion of every instruction stage before the episode limit. Each request derives a noise seed from the task, layout ID, and control step. All methods use this seed for initial future-video latents and the seed plus one for initial action noise.

\begin{table*}[htbp]
\centering
\small
\caption{RoboDojo memory tasks. The limit counts control steps at 25 Hz, not seconds or model requests.}
\label{tab:sim_protocol}
\begin{tabular}{@{}l p{0.66\textwidth} r@{}}
\toprule
Task & Instruction & Limit \\
\midrule
Cover Blocks & Cover the blocks from left to right, remember their colors, then uncover them in the order red, green, blue. & 800 \\
Match and Pick from Conveyor & Remember the first object on the conveyor, then pick the matching object when it appears again. & 700 \\
Press by Number & Press the two red buttons the required number of times according to the number cards, then press the blue button to confirm. & 700 \\
Imitate Sorting Sequence & Observe the object placement order, remember it, then place the corresponding objects into the basket in the same order. & 1600 \\
Swap Blocks & Swap the two blocks using the empty mat, pressing the button after each move. & 700 \\
Swap T & Pick up the two T-shaped blocks, swap their positions, and place them back with the correct orientations. & 400 \\
\bottomrule
\end{tabular}
\end{table*}

\paragraph{Paired outcomes at 30K updates.}
Streaming-history succeeds in 64 of 300 episodes and Full-history in 50. Both solve 40 layouts, whereas 24 are solved only by Streaming-history and 10 only by Full-history. The net gain is concentrated in Cover Blocks (49 versus 35 out of 50). Relative to Window-history, Streaming-history shares 47 solved layouts, gains 17, and loses eight, for nine additional successes overall.

\section{Latency and Memory Measurement Protocol}
\label{app:profiling}
\begin{table*}[htbp]
\centering\small
\caption{Complete-request costs and history intervals, all with RoPE caching. All methods report the median of three per-run statistics, with run ranges in parentheses. Peak allocation covers weights, activations, caches, and workspaces. Cumulative time includes the initial request.}
\label{tab:serving_detail}
\begin{tabular}{@{}lrrr@{}}\toprule
Metric & Streaming-history & Window-history & Full-history \\\midrule
Non-initial P50 (s) & 1.119 (1.099 to 1.170) & 1.492 (1.491 to 1.494) & 4.333 (4.331 to 4.340) \\
Non-initial P95 (s) & 1.124 (1.116 to 1.181) & 1.499 (1.497 to 1.501) & 7.883 (7.878 to 7.887) \\
Initial request (s) & 1.042 (1.036 to 1.115) & 0.967 (0.912 to 0.980) & 0.957 (0.938 to 0.994) \\
Cumulative, 28 requests (s) & 31.1 (30.8 to 32.7) & 40.2 (40.1 to 40.3) & 117.5 (117.4 to 117.6) \\
Complete-request peak (MiB) & 19530.2 & 21286.6 & 21352.0 \\
History of 1 to 10 frames, P50 (s) & 1.096 & 1.324 & 1.327 \\
History of 11 to 30 frames, P50 (s) & 1.117 & 1.490 & 3.150 \\
History of 31 to 52 frames, P50 (s) & 1.121 & 1.497 & 6.524 \\
Video sequence length & $600$ & $1320$ & $360+120n_k$ \\
\bottomrule\end{tabular}
\vspace{1.2em}
\centering\small\setlength{\tabcolsep}{5pt}
\caption{Mean exclusive stage times (ms) with RoPE caching, each over 27 non-initial requests. These separately synchronized diagnostics are distinct from the end-to-end request latency. Inapplicable stages are marked n/a, and 0.0 denotes rounding.}
\label{tab:serving_stages}
\begin{tabular}{@{}lrrr@{}}\toprule
Stage & Streaming-history & Window-history & Full-history \\\midrule
Current observation callback & 4.2 & 4.2 & 4.2 \\
Policy input setup & 0.1 & 0.1 & 0.1 \\
History preparation & 1.6 & 4.3 & 30.9 \\
Current-frame VAE & 11.5 & 11.6 & 11.7 \\
Historical-frame VAE & 22.6 & 69.9 & 254.1 \\
Initial-frame VAE & 0.0 & n/a & n/a \\
GDN writes & 33.5 & n/a & n/a \\
Memory readout and auxiliary tokens & 1.1 & n/a & n/a \\
Text-cache load and transfer & 3.5 & 3.6 & 3.5 \\
VLM input preparation & 9.7 & 9.9 & 10.9 \\
VLM backbone & 34.4 & 34.3 & 37.1 \\
Text context preparation & 0.3 & 0.3 & 0.3 \\
State and mask preparation & 0.2 & 0.3 & 0.3 \\
Remaining condition assembly & 0.2 & 0.1 & 0.1 \\
Latent, noise, and mask setup & 0.6 & 0.4 & 0.4 \\
Joint denoising, ten iterations & 981.1 & 1317.0 & 3984.7 \\
Model return & 0.0 & 0.0 & 0.0 \\
Remaining model call & 0.7 & 0.8 & 0.8 \\
Action readback and interpolation & 0.3 & 0.3 & 0.3 \\
Uninstrumented remainder & 0.2 & 0.2 & 0.2 \\
\midrule
Total profiled request & 1105.9 & 1457.2 & 4339.6 \\
\bottomrule\end{tabular}
\vspace{1.2em}
\centering\small
\caption{Full length-sweep measurements. One Full-history model and the current input are fixed, with three calls per length. Peak increments are relative to the 360-token floor and include historical encoding, activations, and temporary workspaces. The curve appears in main-paper Figure 3b.}
\label{tab:memory_sweep}
\begin{tabular}{@{}rrr@{}}\toprule
History frames & Video tokens & $\Delta M$ (MiB) \\\midrule
0 & 360 & 0.0 \\
2 & 600 & 243.5 \\
8 & 1320 & 1959.4 \\
16 & 2280 & 1965.7 \\
32 & 4200 & 1978.4 \\
48 & 6120 & 1991.0 \\
\bottomrule\end{tabular}\end{table*}

\paragraph{Matched requests.}
We profile one common Imitate Sorting Sequence recording on an exclusively allocated RTX 5090 (1312 frames, with requests at indices $0,48,\ldots,1296$). Each run spans 28 requests, including 27 non-initial requests with 2 to 52 history frames. Inference uses BF16, an FP32 memory state, ten denoising iterations, the library's recurrent kernel, and one CPU/OMP/MKL/OpenBLAS thread under PyTorch 2.11.0+cu128 and CUDA 12.8.

Each request is timed from the current-observation callback through history maintenance and action generation until the $48\times14$ control targets return to the CPU, synchronizing CUDA only at these boundaries. Model loading, a two-request warmup followed by reset, input decoding, and intervening observation callbacks are excluded. Cumulative time sums inference requests on the fixed recording.

All reported latencies apply RoPE caching across three repeats per method, evaluated in separate sessions on matched inputs. We compute statistics within each repeat and report the median of three per-run statistics, using linear interpolation at $(N-1)q$ for quantiles. Speedups pair matching request indices within each comparison and take the median across the per-repeat paired medians.

\paragraph{Complete serving costs.}
Table~\ref{tab:serving_detail} lists request latencies, cumulative costs, history-interval P50s (over 5, 10, and 12 requests), peak allocated memory, and video sequence lengths. Window-history conditions on the initial frame plus up to seven recent observations, deduplicated to at most eight frames. Table~\ref{tab:serving_stages} breaks down mean stage times, where Full-history historical VAE encoding approaches 480\,ms near the end of the recording. Per-run linear fits yield a median latency slope of 0.26\,ms per history frame for Streaming-history, versus 4.4\,ms for the filling window and 149.8\,ms for Full-history.

\paragraph{Sequence length and memory.}
To isolate sequence-length cost at a fixed observation resolution, we sweep the historical prefix of one Full-history model with fixed current input (\cref{tab:memory_sweep}). We define $\Delta M(L)=M_{\mathrm{peak}}(L)-M_{\mathrm{peak}}(360)$ over a 19377.46\,MiB floor. The compact sequence reduces the allocation increment, whereas the explicit-history peak largely plateaus beyond eight frames under chunked encoding and fused attention while Full-history latency continues to grow.

Whereas Table~\ref{tab:serving_detail} reports total serving peaks, persistent representation storage is much smaller. The memory state occupies 1\,MiB, and each BF16 historical latent occupies $48\times24\times20\times2=46080$ bytes (45\,KiB). A full window holds at most 360\,KiB of latents, and retaining all sampled latents takes $45n_k$\,KiB, excluding model weights, temporary activations, and unwritten RGB observations.

\paragraph{Incremental updates.}
We compare incremental updates against prefix reconstruction on three 28-request Imitate Sorting Sequence recordings using the Streaming-history checkpoint with RoPE caching. Both variants share the checkpoint, matched inputs, noise seeds, recurrent kernel, historical-latent cache, and preparation settings, and both encode only newly added frames. Incremental updates project and write each frame once, whereas prefix reconstruction re-projects and rewrites the full prefix from an empty state. All 84 action hashes and all 81 non-initial final state hashes match pairwise, and the actions also match the uncached control. Across the 81 non-initial requests, incremental updates yield a $1.392\times$ median paired speedup (per-trajectory medians of 1.381, 1.392, and 1.402), cutting writes from 738 to 52 per trajectory and reducing complete trajectory time from about 45\,s to about 32\,s.

\paragraph{RoPE caching.}
Within each request, our serving path caches rotary position embedding (RoPE) tables and device-resident frequency tensors for reuse across layers and denoising iterations, while still recomputing layerwise hidden states and attention. RoPE caching is applied to all methods. Comparing cached and uncached execution, all 84 Streaming-history, 28 Window-history, and 84 Full-history action pairs have identical hashes, and all 140 action checks between profiling and timing runs agree. This supports using the cached path for serving costs while retaining the original closed-loop evaluations. Peak allocation is the same with and without caching.
\section{Real-Robot Experiments}
\label{app:robot}
\suppressfloats[t]

\begin{figure}[tbp]
\centering
\includegraphics[width=\textwidth]{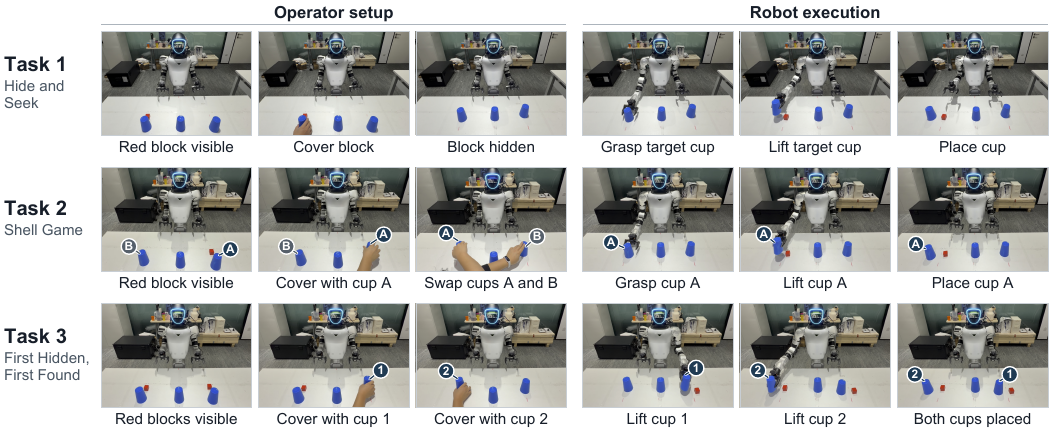}
\caption{\textbf{Real-robot task overview.} Key frames show operator setup and robot execution for each task. In Shell Game, A and B identify the exchanged cups. In First Hidden, First Found, 1 and 2 mark the order in which the target cups cover the blocks and are lifted. The robot execution frames come from Streaming-history trials.}
\label{fig:real_robot_tasks}
\end{figure}

We evaluate whether streaming visual memory supports physical manipulation on a Unitree G1-D dual-arm robot when task evidence appears mid-episode and must be retained until execution. Across one tabletop setup with three identical blue cups and red blocks, every episode ends with the robot lifting a covering cup and placing it to the right of its block. Task 1 (Hide and Seek) is the main task, accompanied by three test variants that alter when evidence is used, how the arm begins its reach, and whether the evidence is later revised. Tasks 2 and 3 change how the blocks are hidden (Figure~\ref{fig:real_robot_tasks}).

\begin{table}[htbp]
\centering\small
\caption{Successful trials out of 20 for each condition and method. Indented rows are test variants of Hide and Seek, evaluated with the Hide and Seek checkpoints.}
\label{tab:real_robot}
\begin{tabular}{@{}lcccc@{}}
\toprule
Condition & No-history & Window-history & Full-history & Streaming-history (ours) \\
\midrule
Task 1 (Hide and Seek) & 6 & 16 & \textbf{20} & 19 \\
\quad Wait, Then Seek & 7 & 8 & 14 & \textbf{20} \\
\quad Pushed to a Wrong Cup & 0 & 12 & 14 & \textbf{19} \\
\quad Hidden Twice & 4 & 11 & 8 & \textbf{20} \\
Task 2 (Shell Game) & 6 & 17 & \textbf{18} & \textbf{18} \\
Task 3 (First Hidden, First Found) & 7 & 19 & \textbf{20} & \textbf{20} \\
\midrule
All six conditions (out of 120) & 30 & 83 & 94 & \textbf{116} \\
\bottomrule
\end{tabular}
\end{table}
Table~\ref{tab:real_robot} summarizes all six conditions. On the three base tasks, all three history representations succeed in at least 16 of 20 trials, whereas No-history reaches at most 7. The Task 1 test variants separate the history representations, with Streaming-history succeeding in 59 of 60 trials compared with 36 for Full-history and 31 for Window-history.

\subsection{Setup}

\paragraph{Robot and inputs.}
The policy observes the left high camera stacked above the left and right wrist cameras on a $384\times320$ canvas, matching simulation. It runs at 30\,Hz and predicts 64 actions per request. History frames are sampled at 1\,Hz for Task 1 and its variants (frames 0, 30, 60, and so on) and at 2\,Hz for Tasks 2 and 3 to record cup exchanges and covering sequences more densely. Spatial directions such as right refer to the robot frame.

\paragraph{Methods and training.}
The four methods match the main paper. Window-history supplies the initial frame and up to seven recent sampled observations, covering 7\,s at 1\,Hz and 3.5\,s at 2\,Hz, whereas Full-history supplies all prior sampled observations and Streaming-history uses $C_k$. For each task, all methods are fine-tuned from the pretrained EWAM checkpoint on the same 100 demonstrations for 5000 updates.

\paragraph{Evaluation.}
All methods share 20 fixed layouts per condition, where each layout specifies the randomized block placement and operator move sequence. A trial succeeds only if the robot completes the entire instruction, including placing every lifted cup. In Tasks 1 to 3, the robot starts within about 1\,s of the last operator action. For Task 1 and its variants, we categorize failures as grasp failures when the robot targets the correct cup and wrong selections otherwise.

\subsection{Task 1, Hide and Seek}

\begin{figure}[htbp]
\centering
\includegraphics[width=\textwidth]{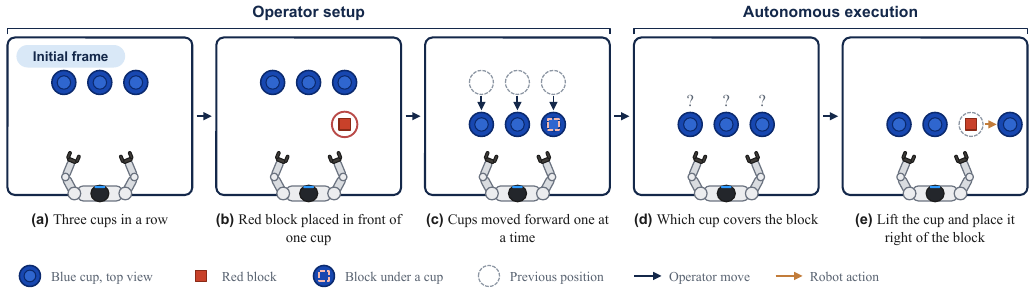}
\caption{\textbf{Hide and Seek.} The red block appears after the initial frame, and the operator moves the cups forward one at a time until one of them covers it. The robot lifts the covering cup and places it to the right of the block. The target position and the pushing order are randomized.}
\label{fig:robot_hide}

\vspace{1em}
{\small
\captionof{table}{Outcomes of Hide and Seek, out of 20 trials per method.}
\label{tab:robot_hide}
\begin{tabular}{@{}lccc@{}}
\toprule
Method & Success & Grasp failure & Wrong selection \\
\midrule
No-history & 6 & 0 & 14 \\
Window-history & 16 & 3 & 1 \\
Full-history & 20 & 0 & 0 \\
Streaming-history (ours) & 19 & 1 & 0 \\
\bottomrule
\end{tabular}
}
\end{figure}

\paragraph{What it tests.}
The instruction is ``Pick up the cup covering the red block and place it to the right of the block.'' Because the block appears only after the initial frame and the operator moves all three identical cups forward in random order, neither the initial frame nor the current observation identifies the hidden target. The policy must retain which cup covered the block during setup rather than simply picking the last cup moved.

\paragraph{Results.}
When the robot acts immediately after setup, all three history representations identify the target cup in nearly every trial (Table~\ref{tab:robot_hide}). Full-history completes all 20 trials, and Streaming-history reaches the target cup in all 20 with a single grasp failure. No-history fails by wrong selection in all 14 of its failures, and its 6 successes are close to the rate of choosing among three identical cups at random.

\subsection{Test variants of Task 1}

Three unseen test variants reuse the scene, instruction, and checkpoints of Task 1 while altering a single episode factor to probe out-of-distribution histories.

\begin{figure}[htbp]
\centering
\includegraphics[width=\textwidth]{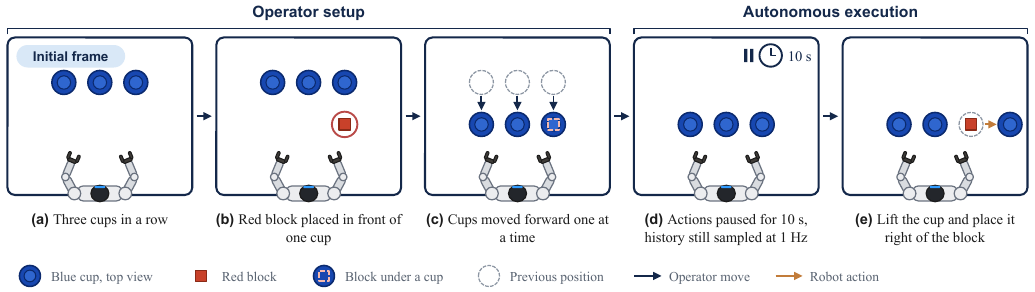}
\caption{\textbf{Wait, Then Seek.} After the cups cover the block, the actions of the robot are paused for 10\,s while history frames are still sampled at 1\,Hz. The robot then lifts the covering cup.}
\label{fig:robot_wait}

\vspace{1em}
{\small
\captionof{table}{Outcomes of Wait, Then Seek, out of 20 trials per method.}
\label{tab:robot_wait}
\begin{tabular}{@{}lccc@{}}
\toprule
Method & Success & Grasp failure & Wrong selection \\
\midrule
No-history & 7 & 0 & 13 \\
Window-history & 8 & 0 & 12 \\
Full-history & 14 & 6 & 0 \\
Streaming-history (ours) & 20 & 0 & 0 \\
\bottomrule
\end{tabular}
}
\end{figure}

\paragraph{Wait, Then Seek.}
Pausing for 10\,s adds about ten static frames of the covered scene, pushing the covering event past the seven-frame capacity of Window-history. Window-history consequently drops to 8 successes (Table~\ref{tab:robot_wait}), moving directly to a wrong cup in all 12 failures. Full-history retains the covering event and targets the correct cup in all 20 trials, but loses six trials to premature or late gripper closure (Figure~\ref{fig:robot_cases}d). Because the robot executes each 64-action chunk in about 2.1\,s at 30\,Hz and Full-history takes a median of at least 3.15\,s per request after the pause (Table~\ref{tab:serving_detail}, measured on the same type of GPU), the arm waits longer between chunks than it moves, whereas Window-history and Streaming-history wait about 1.5\,s and 1.1\,s. We attribute the grasp failures of Full-history to these long waits. Streaming-history succeeds in all 20 trials, confirming that evidence written before the pause remains usable afterward.

\begin{figure}[htbp]
\centering
\includegraphics[width=\textwidth]{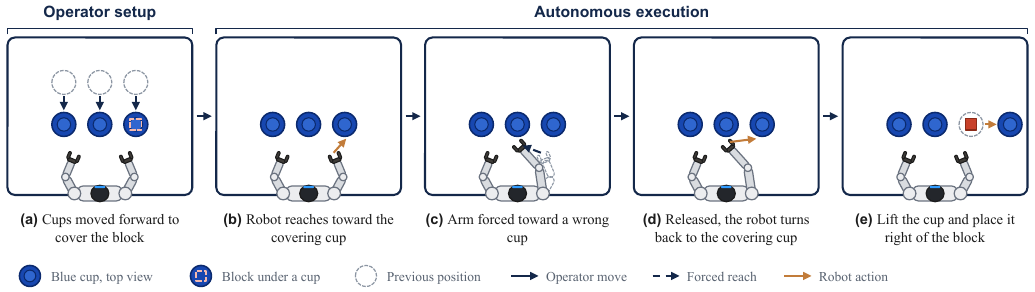}
\caption{\textbf{Pushed to a Wrong Cup.} Before or during the reach, the operator drives the arm toward a wrong cup and then releases it. The robot has to return to the covering cup and complete the placement.}
\label{fig:robot_push}

\vspace{1em}
{\small
\captionof{table}{Outcomes of Pushed to a Wrong Cup, out of 20 trials per method.}
\label{tab:robot_push}
\begin{tabular}{@{}lccc@{}}
\toprule
Method & Success & Grasp failure & Wrong selection \\
\midrule
No-history & 0 & 0 & 20 \\
Window-history & 12 & 0 & 8 \\
Full-history & 14 & 2 & 4 \\
Streaming-history (ours) & 19 & 1 & 0 \\
\bottomrule
\end{tabular}
}
\end{figure}

\paragraph{Pushed to a Wrong Cup.}
Displacing the arm places it beside a wrong cup at release, so the current observation and recent frames conflict with the earlier covering event. Table~\ref{tab:robot_push} lists the outcomes. Every wrong selection by Window-history (8 trials) and Full-history (4 trials), along with all 20 No-history trials, lifts the wrong cup beside the displaced arm. When Streaming-history and Full-history succeed, the arm redirects straight to the correct cup after release without pausing or touching the wrong cup. Streaming-history identifies the correct cup in all 20 trials, with one grasp failure.

\begin{figure}[htbp]
\centering
\includegraphics[width=\textwidth]{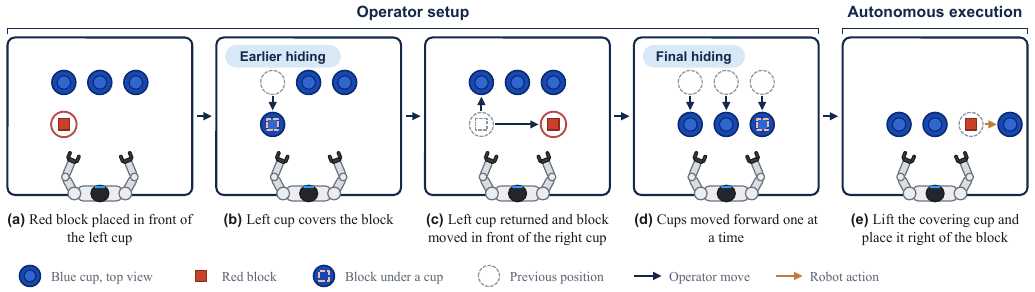}
\caption{\textbf{Hidden Twice.} The block is hidden under one cup, revealed, moved in front of another cup, and hidden again as the cups move forward. Only the final hiding determines the target.}
\label{fig:robot_twice}

\vspace{1em}
{\small
\captionof{table}{Outcomes of Hidden Twice, out of 20 trials per method.}
\label{tab:robot_twice}
\begin{tabular}{@{}lccc@{}}
\toprule
Method & Success & Grasp failure & Wrong selection \\
\midrule
No-history & 4 & 0 & 16 \\
Window-history & 11 & 0 & 9 \\
Full-history & 8 & 0 & 12 \\
Streaming-history (ours) & 20 & 0 & 0 \\
\bottomrule
\end{tabular}
}
\end{figure}

\paragraph{Hidden Twice.}
When the block is revealed and hidden under a second cup, both hiding events remain in the history, but the first is outdated. All 12 failures of Full-history and all 9 of Window-history lift the first cup, showing that both explicit representations act on stale evidence, whereas Streaming-history follows the revised hiding in all 20 trials (Table~\ref{tab:robot_twice}).

\paragraph{Summary of Task 1.}
Across all 80 trials of Task 1 and its variants, Streaming-history selects the correct cup in all 80 trials (failing twice during grasping), compared with 64 for Full-history, 50 for Window-history, and 17 for No-history. Each explicit representation fails in ways tied to the perturbation (Figure~\ref{fig:robot_cases}). Window-history loses the target after the 10\,s pause, whereas both explicit representations are misled by the displaced arm and by the outdated first hiding.

\providecommand{\TaskOneFailureFigure}{realrobot_task1_failures_with_grasp}
\begin{figure}[tbp]
\centering
\includegraphics[width=\textwidth]{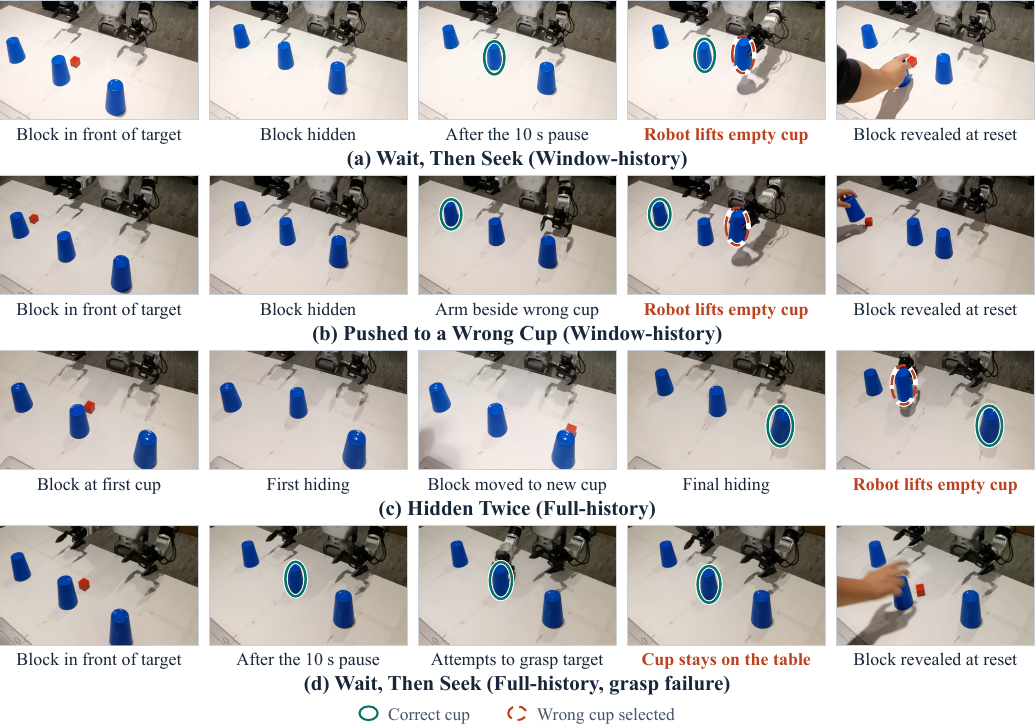}
\caption{\textbf{Failure cases in the Task 1 variants.} Each row shows one trial filmed from an external camera. (a) Window-history lifts an empty cup after the 10\,s pause. (b) Window-history lifts the empty cup beside its displaced arm. (c) Full-history lifts the cup from the first hiding after the block is moved and hidden again. (d) Full-history selects the correct cup after the 10\,s pause but fails to lift it. Solid outlines mark the correct cup, and dashed outlines mark a wrong cup that the robot lifts. In (a), (b), and (d), the operator reveals the block during reset.}
\label{fig:robot_cases}
\end{figure}

\subsection{Tasks 2 and 3}

\begin{figure}[htbp]
\centering
\includegraphics[width=\textwidth]{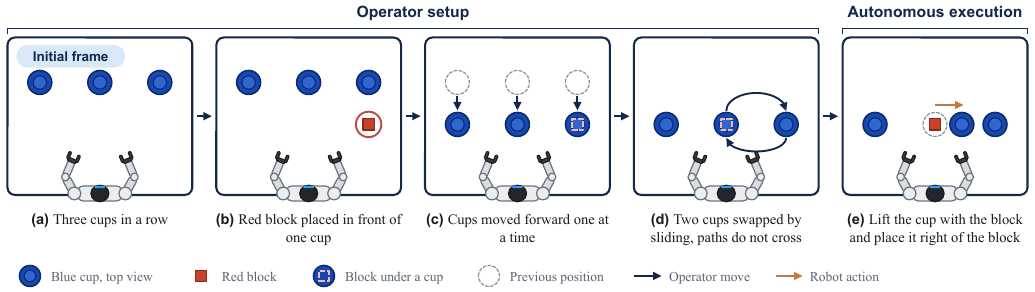}
\caption{\textbf{Shell Game.} After the block is covered, the operator swaps two cups by sliding them along paths that do not cross. The robot lifts the cup that now covers the block.}
\label{fig:robot_shell}
\end{figure}

\paragraph{Task 2, Shell Game.}
Using the Task 1 instruction, the operator swaps two cups after covering the block, requiring the policy to track the moving cup whenever the target is exchanged. All three history representations reach 17 or 18 successes (Table~\ref{tab:real_robot}). Every failure of the three history representations lifts the cup resting at the block's original location rather than tracking the swap.

\begin{figure}[htbp]
\centering
\includegraphics[width=\textwidth]{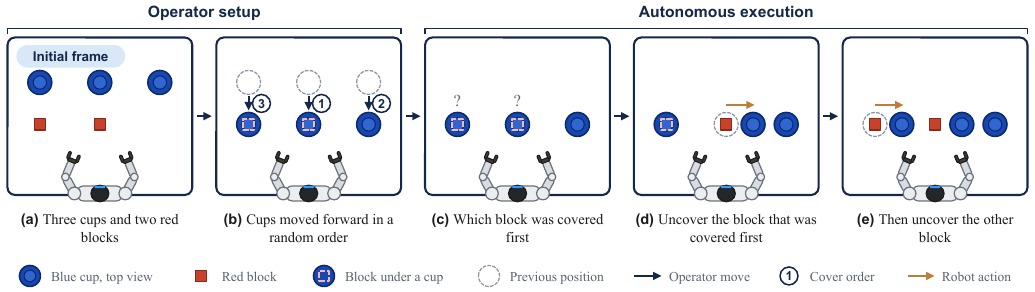}
\caption{\textbf{First Hidden, First Found.} Two red blocks are visible in the initial frame. The operator moves the cups forward in a random order, so the empty position can be covered between the two blocks. The robot uncovers the blocks in the order in which they were covered and places each cup to the right of its block.}
\label{fig:robot_order}
\end{figure}

\paragraph{Task 3, First Hidden, First Found.}
The instruction is ``Identify the cups covering the red blocks. Pick up the cups in the order in which they covered the blocks, placing each cup to the right of its corresponding block.'' Although block positions appear in the initial frame, the covering order is available only from history, with the empty cup's movement acting as a distractor. Streaming-history and Full-history succeed in all 20 trials and Window-history in 19 (Table~\ref{tab:real_robot}), with the lone Window-history failure occurring during manipulation rather than ordering.

\paragraph{What Tasks 2 and 3 show.}
Tasks 2 and 3 test whether fixed-size memory preserves cup motion and multi-event ordering. Because execution starts within about 1\,s of the final operator move, the relevant events remain inside the recent window, allowing all three history representations to succeed in at least 17 of 20 trials and within one trial of each other. Streaming-history matches Full-history on both tasks using a fixed token count instead of reprocessing every sampled frame at each request. Combined with the Task 1 variants, where evidence falls outside the window, conflicts with the current arm pose, or is later revised, these results suggest that the recurrent state preserves motion and order cues while its advantage appears when evidence must persist or be updated.

\end{document}